\documentclass[11pt]{article}

\usepackage[final]{acl}

\usepackage{amsmath}
\usepackage{booktabs}
\usepackage{times}
\usepackage{latexsym}

\usepackage{soul}
\usepackage{xcolor}
\sethlcolor{yellow}

\usepackage[T1]{fontenc}

\usepackage[utf8]{inputenc}

\usepackage{microtype}

\usepackage{inconsolata}

\usepackage{graphicx}

\title{Large Language Models as Unified Multimodal Learners for Clinical Prediction}

\author{
Ajay Madhavan Ravichandran$^{1}$ \qquad
Bilgin Osmandoja$^{2}$ \qquad
Klemens Budde$^{2}$ \\
\textbf{Klaus Netter}$^{3}$ \qquad
\textbf{Tobias Strapatsas}$^{4}$ \qquad
\textbf{Aljoscha Burchardt}$^{1}$ \\
\textbf{Sebastian Möller}$^{1,5}$ \qquad
\textbf{Roland Roller}$^{1,2,5}$ \\
$^{1}$German Research Center for Artificial Intelligence (DFKI), Germany\\
$^{2}$Charité Universitätsmedizin Berlin, Germany, 
$^{3}$DNC Information Management GmbH, Germany\\
$^{4}$Klinik für Akut- und Notfallmedizin, Asklepios Klinikum Harburg, Germany\\
$^{5}$Technical University Berlin, Germany
}

\begin{document}
\maketitle
\begin{abstract}
Electronic health records combine free-text clinical narratives with structured measurements such as vital signs, laboratory values, and comorbidities. Yet most clinical prediction systems still rely on task-specific fusion architectures, pairing dedicated encoders for each modality with learned combination mechanisms that must be re-engineered for every new task and clinical setting. We propose a simpler alternative: convert all patient data, regardless of modality, into a single natural language sequence and fine-tune a pretrained language model end-to-end, with no architectural modification for fusion. We evaluate this approach across three clinically distinct prediction tasks — in-hospital mortality on MIMIC-III, graft failure prediction using longitudinal data from a German transplant center, and emergency triage classification from ambulance records — comparing encoder-based (ModernBERT) and decoder-based (Llama 3.1, Gemma, DeepSeek-R1-Qwen, Qwen3) fine-tuning against established multimodal baselines and, for graft failure, a gradient boosting model currently used in clinical practice for post-transplant patient management. Across all three tasks, unified textual serialization matches or exceeds task-specific multimodal baselines, and outperforms the clinically deployed gradient boosting system on graft failure prediction. These results indicate that a single serialization-based paradigm, without bespoke fusion architectures, is sufficient for multimodal clinical prediction — substantially reducing system complexity while matching or exceeding specialized designs.
\end{abstract}

\section{Introduction}

Electronic health records (EHRs) are inherently multimodal: they capture patient information across fundamentally different representational spaces, combining free-text clinical narratives authored by physicians with structured measurements such as laboratory values, vital signs, demographics, and coded diagnoses \cite{johnson2016mimic}. Effective clinical prediction requires integrating these modalities; however, they are not easily combined. The prevailing approach has been to design task-specific fusion architectures, consisting of dedicated encoders for each modality followed by learned combination mechanisms such as gated attention or cross-modal transformers \cite{khadanga2019using, soenksen2022integrated, cui2024hypergraph, kim2026lemof}. While effective, this paradigm carries significant practical limitations: each new clinical task requires a custom architecture, modality-specific preprocessing pipelines, and extensive tuning. As a result, existing systems are often fragmented, difficult to reuse, and costly to deploy across diverse clinical settings \cite{benmiled2025scoping}.
 
Large Language Models (LLMs) have recently emerged as a potential alternative. A growing body of work shows that structured data can be serialized into natural language and processed by pretrained LLMs, removing the need for modality-specific encoders. CPLLM ~\cite{cpllm2024} demonstrated that fine-tuning Llama2 on text-serialized EHR records outperforms Med-BERT on disease prediction and readmission tasks without domain-specific pretraining. Hegselmann et al.~\cite{hegselmann2025ehr} further showed that LLM embeddings of serialized structured patient records match or exceed specialized EHR foundation models across multiple clinical tasks. These studies establish that serialization is effective for structured data in isolation. However, an important limitation remains: \textbf{real-world EHR prediction tasks are inherently multimodal}. Clinical decisions depend on both physician-authored narratives and structured measurements, which provide complementary information \cite{ruan2025evidence}. Existing serialization-based approaches have not systematically addressed this setting, where both modalities must be integrated within a unified framework.
 
In this work, we address this gap by proposing a unified framework in which \emph{all} patient data, regardless of modality, is converted into a single natural language sequence and used to fine-tune a pretrained language model end-to-end. Structured EHR variables—including demographics, comorbidities, laboratory results, and vital signs—are serialized into key--value textual representations and concatenated with clinical narratives. This unified representation enables the model to learn cross-modal relationships through its attention mechanism without requiring architectural modifications. Importantly, the same model architecture and fine-tuning procedure can be applied across different combinations of modalities and prediction tasks, avoiding the need for task-specific fusion designs.
 
We evaluate this approach across three clinical prediction tasks with distinct data sources and objectives: in-hospital mortality prediction on MIMIC-III~\cite{johnson2016mimic}, combining clinical notes with temporal physiological features; graft failure prediction using data from a German transplant center, combining clinical narratives with longitudinal laboratory and comorbidity data \cite{roller2022evaluation}; and emergency triage classification, combining short ambulance notes with structured vital measurements and neurological scores \cite{maschhur2024towards}. For each task, we compare encoder-based fine-tuning (ModernBERT~\cite{warner2025smarter}) and decoder-based supervised fine-tuning (Llama 3.1~\cite{grattafiori2024llama}, Gemma~\cite{team2024gemma}, DeepSeek-R1-Qwen~\cite{deepseekai2025deepseekr1}, and Qwen3~\cite{qwen3technicalreport}) against established unimodal and multimodal baselines, as well as a gradient boosting system currently used in clinical practice for graft failure prediction.
 
Our results show that the proposed serialization approach achieves performance comparable to or exceeding task-specific multimodal baselines across all tasks. These findings suggest that a unified text-based representation can serve as an effective alternative to specialized multimodal architectures, while substantially reducing system complexity.
 
The contributions of this work are as follows:
 
\begin{itemize}
    \item \textbf{Unified multimodal serialization for EHR prediction.} We propose a general framework that converts heterogeneous EHR data—including clinical narratives and structured variables—into a single textual representation compatible with pretrained language models, eliminating the need for task-specific architectures.
 
    \item \textbf{Systematic evaluation across models and domains.} We compare encoder-based and decoder-based LLM fine-tuning across three clinical prediction tasks spanning critical care, transplant medicine, and emergency triage, demonstrating the generality of the proposed approach.
 
   \item \textbf{Comparison with clinical baselines.} We show that the proposed method matches or exceeds established research baselines and outperforms a gradient boosting system currently deployed in clinical practice for graft failure prediction, underscoring its potential for real-world clinical use.
\end{itemize}
\section{Related Work}
\label{sec:related}
\subsection{Multimodal Fusion Architectures for EHR Prediction}
The dominant paradigm for clinical prediction from heterogeneous EHR data is to process each modality with a dedicated encoder and combine representations through explicit fusion mechanisms. Rajkomar et al.~\cite{rajkomar2018scalable} showed that deep learning over heterogeneous EHR data can outperform traditional statistical models, and Khadanga et al.~\cite{khadanga2019using} established a widely used baseline combining clinical notes with ICU time-series via separate encoders for mortality prediction on MIMIC-III. Soenksen et al.~\cite{soenksen2022integrated} generalized this with HAIM, fusing imaging, tabular, time-series, and text modalities through late fusion.
More recent work has pursued increasingly sophisticated fusion strategies: hypergraph-based fusion of structured data with LLM-derived text embeddings~\cite{cui2024hypergraph}, evidence-based fusion grounded in belief function theory~\cite{ruan2025evidence}, multi-agent architectures where modality-specific LLM agents summarize inputs before prediction~\cite{gao2025moma}, and layer-level fusion across modality-specific encoders~\cite{kim2026lemof}. Closer to our approach, EHR2Path~\cite{2026ehr2path} converts longitudinal records into a unified textual representation, but still relies on a purpose-built summarization module to compress long patient histories and targets trajectory simulation rather than direct outcome prediction.
Despite strong performance, these approaches share a common limitation: even the most generalizable, such as HAIM and MoMA, still require a modality-specific encoder, agent, or compression mechanism to be designed for each new data type, demanding substantial engineering effort that does not readily generalize across tasks. A recent scoping review~\cite{benmiled2025scoping} highlights the handcrafted nature of most fusion architectures as a major barrier to scalable clinical deployment.
\subsection{LLM-Based Serialization for Structured and Clinical Data}
Pretrained language models have advanced clinical NLP through domain-adapted encoders such as BioClinicalBERT~\cite{alsentzer2019publicly}, BioBERT~\cite{lee2020biobert}, and GatorTron~\cite{yang2022gatortron}, and more recently through large-scale generative models: Med-PaLM~\cite{singhal2023large} achieved near-clinician performance on medical examinations, and GPT-4 has shown strong results on medical challenge problems~\cite{nori2023capabilities}.
Building on these advances, a growing line of work fine-tunes LLMs directly on serialized structured data. CPLLM~\cite{cpllm2024} fine-tuned Llama2 on text-serialized EHR codes and outperformed Med-BERT on disease and readmission prediction without domain-specific pretraining, and Hegselmann et al.~\cite{hegselmann2025ehr} showed that LLM embeddings of serialized structured records match or exceed specialized EHR foundation models across multiple tasks. Outside the clinical domain, LIFT~\cite{dinh2022lift} and TabLLM~\cite{hegselmann2023tabllm} showed that fine-tuning LLMs on serialized tabular rows can match gradient-boosted models on general tabular classification and regression, and a recent survey by Fang et al.~\cite{fang2024survey} synthesizes the effectiveness of serialization-based approaches across prediction, generation, and understanding tasks on tabular data. In the emergency care setting, Lee et al.~\cite{edpseudo2024} introduced MEME, which converts multimodal tabular EHR data into synthetic clinical pseudo-notes and outperformed traditional machine learning, EHR foundation models, and GPT-4 prompting on emergency department decision support tasks. Related work has further extended LLM fine-tuning to longitudinal and temporal clinical reasoning: DT-GPT~\cite{dtgpt2025} fine-tunes LLMs to forecast clinical variable trajectories without architectural modification, TIMER~\cite{cui2025timer} instruction-tunes LLMs for temporal reasoning across multi-visit records, and Wu et al.~\cite{zhen2024} introduced instruction tuning for open-ended EHR question answering using the MIMIC-Instr dataset.

Together, these studies establish that serialization is an effective interface between structured or code-based EHR data and pretrained language models, and that fine-tuning on serialized inputs can match or exceed specialized baselines without domain-specific pretraining or architectural changes. However, in each case the model is trained on a single source of information -- structured codes, tabular features, or generated instruction-response pairs -- rather than the combination of free-text clinical narratives and structured measurements that characterizes real-world EHR prediction tasks.
At the same time, recent evidence suggests that this gap cannot simply be closed by prompting general-purpose LLMs. El Khettari et al.~\cite{elkhettari2026clinical} found that supervised multimodal fusion of clinical text and structured variables outperformed LLM-based approaches, which performed inconsistently across modalities and decoding strategies. Yildiz et al.~\cite{yildiz2025transform} further highlight open methodological challenges for LLMs in clinical prediction, including time-to-event modelling, calibration, and external validation. Together, these findings motivate our focus on supervised fine-tuning of LLMs on unified textual serializations, rather than zero-shot or few-shot prompting, as the more promising path toward closing the multimodal gap.
In contrast to all of the above, our work fine-tunes pretrained LLMs on a single serialized sequence that jointly encodes both structured and unstructured patient information, and evaluates this approach directly against task-specific multimodal fusion baselines and a clinical gradient boosting system across three distinct prediction tasks.
\begin{table*}[!t]
\small
\begin{center}
\begin{minipage}{\textwidth}
\caption{Overview of the datasets, input modalities, and prediction tasks used in the evaluation. \label{tab:datasets}}%
\begin{tabular*}{\textwidth}{@{\extracolsep{\fill}}llll@{}}
\toprule
Dataset & Input Modalities & Target Task & Evaluation Metrics \\
\midrule
MIMIC-III & Clinical Notes + 13 Time-series features & In-hospital Mortality (48h) & AUROC, AUPRC \\
Graft Failure & Clinical Texts + Labs + Vitals + Comorbidities & Graft Failure (360 days) & AUROC, AUPRC \\
Triage & Short Notes + Vitals + Pain Score + GCS & MTS Urgency (Multi-class) & F1, Precision, Recall \\
\bottomrule
\end{tabular*}
\end{minipage}
\end{center}
\end{table*}

\section{Datasets and Tasks}
\label{sec:datasets}
In this study, we evaluate our proposed methodology across three distinct clinical prediction tasks. Each dataset presents a unique challenge in multimodal learning, requiring the synthesis of unstructured clinical narratives with structured electronic health record (EHR) data. The selected tasks cover various clinical settings, including intensive care, long-term outpatient monitoring, and emergency response.
\subsection{In-Hospital Mortality Prediction (MIMIC-III)}
The first task involves predicting in-hospital mortality among Intensive Care Unit (ICU) patients, utilizing the publicly available MIMIC-III database. This is formulated as a binary classification task to predict patient mortality during the hospital stay, conditioned strictly on data collected within the first 48 hours of ICU admission. The feature space fuses unstructured clinical admission notes with 13 structured temporal vital signs and laboratory results. To ensure numerical stability, temporal measurements undergo strict standardization, including unit normalization, outlier filtering, and hourly aggregation. The dataset comprises a large-scale cohort of 21,139 ICU admissions. Following established benchmarks, the data is partitioned into a training set of 14,681 admissions (approx.\ 70\%), a validation set of 3,222 admissions (15\%), and a held-out test set of 3,236 admissions (15\%). This yields a rich corpus of over 1.2 million distinct temporal data points alongside corresponding clinical notes. Performance is assessed using the Area Under the Receiver Operating Characteristic curve (AUROC) and the Area Under the Precision-Recall Curve (AUPRC), the latter being particularly critical due to the inherent class imbalance of mortality events.

\subsection{Graft Failure Prediction}
Sourced from a comprehensive clinical decision support system study at a German transplant center, the second dataset focuses on the longitudinal monitoring of kidney transplant recipients. The primary objective is a binary classification task aimed at predicting the occurrence of death-censored graft failure within 360 days following a specific outpatient clinical visit. The model processes unstructured clinical progress notes in tandem with a robust set of structured variables, including patient demographics, documented comorbidities, and longitudinal trajectories of vital signs and laboratory results. The cohort encompasses 1,516 kidney transplant recipients, capturing over 100,000 distinct routine clinical data points collected during post-operative care. To simulate real-world prospective deployment, we employ a chronological split, allocating 70\% of the longitudinal records for training, 10\% for hyperparameter tuning, and the final 20\% for independent testing. Consistent with the mortality task, predictive performance is evaluated using AUROC and AUPRC to account for the skewed, rare distribution of graft failure occurrences.

\subsection{Emergency Triage Classification (MTS)}
The final dataset consists of semi-structured ambulance records and triage assignments from a German emergency department, presenting a high-pressure, time-sensitive clinical scenario. This is framed as a multi-class classification problem designed to predict the patient's treatment urgency level upon emergency department arrival, adhering to the Manchester Triage System (MTS) guidelines. Features include short-form, often noisy text narratives describing the immediate clinical situation, coupled with structured physiological markers such as vital signs, pain scores, and the Glasgow Coma Scale (GCS). The dataset consists of over 18,000 anonymized emergency ambulance records collected over a two-year period. The dataset is partitioned into an 80\% training split (approx.\ 14,400 records), a 10\% validation split, and a 10\% test split, strictly stratified by MTS triage urgency levels to preserve class distributions. Due to the multi-class nature of the problem, the model is evaluated using Precision, Recall, and the Macro-averaged F1-score to equally weight the predictive performance across all triage categories, regardless of their frequency.

\section{Method}
\label{sec:method}

\subsection{Problem Formulation}

We consider clinical prediction tasks where each patient encounter consists of both unstructured clinical narratives and structured electronic health record (EHR) data. The structured data includes variables such as vital signs, laboratory measurements, demographics, and comorbidities.

Formally, each sample can be represented as a pair:

\[
(x_{text}, x_{struct})
\]

where $x_{text}$ denotes the clinical narrative and $x_{struct}$ represents the set of structured features associated with the patient encounter. The learning objective is to predict a clinical outcome $y$, which may correspond to either a binary classification task (e.g., mortality prediction) or a multi-class classification task (e.g., triage level).

Traditional multimodal approaches process these modalities using separate encoders and combine their representations through fusion mechanisms. In contrast, our approach converts structured data into a textual representation, allowing a single language model to process both modalities jointly.

\subsection{Unified Multimodal Text Serialization}

To enable language models to process heterogeneous clinical inputs, we convert all structured variables into a textual format and concatenate them with the clinical narrative. This procedure produces a unified sequence that can be directly processed by a pretrained language model.

For each patient record, structured variables are serialized as key–value pairs describing the clinical measurements. An example serialized input is shown in Figure \ref{fig:example1} and Figure \ref{fig:example2}.

\begin{figure}[ht]
    \centering
    \fbox{\parbox{0.47\textwidth}{
\textbf{Patient Information}:\\
\textbf{Age}: 67\\
\textbf{Sex}: Male\\
\textbf{Heart Rate}: 92 bpm\\
\textbf{Blood Pressure}: 130/80 mmHg\\
\textbf{Creatinine}: 1.4 mg/dL\\
\textbf{Pain Score}: 6\\
\\
\textbf{Clinical Note}:\\
A 67-year-old male presented to the emergency department
complaining of substernal chest pain radiating to the left
shoulder. Symptoms began approximately two hours prior to
admission and were accompanied by shortness of breath and
mild diaphoresis. The patient reports a history of
hypertension and hyperlipidemia and is currently taking
amlodipine and atorvastatin...
    }}
    \caption{Example of a serialized input with text and time invariant data}
    \label{fig:example1}
\end{figure}

\begin{figure}[ht]
    \centering
    \small
    \fbox{\parbox{0.48\textwidth}{
\textbf{Patient Information}:\\
\textbf{Heart Rate}: 76.09, 78.75, 76.88, 69.75\\
\textbf{Respiratory Rate}: 19.19, 16.62, 16.62, 17.5\\
\textbf{Systolic Blood Pressure}: 136.71, 129.94, 140.71, 144.56\\
\textbf{Diastolic Blood Pressure}: 65.25, 56.06, 59.69, 55.81\\
\textbf{Mean Blood Pressure}: 85.62, 76.5, 80.42, 78.5\\
\textbf{Oxygen Saturation}: 97.08, 96.38, 96.62, 96.38\\
\textbf{Temperature}: 36.68, 36.56, 37.06, 37.0\\
\textbf{Glucose}: 165.0, 127.0, 128.0, 128.0\\
\textbf{Glasgow Coma Scale Total}: 11.0, 11.0, 11.0, 11.0\\
\textbf{PH}: 7.4, 7.4, 7.4, 7.4\\
\textbf{Fraction Inspired Oxygen}: 0.21, 0.21, 0.21, 0.21\\
\textbf{Weight}: 90.0\\
\textbf{Height}: 170.0 \\
\\
\textbf{Clinical Note}:\\
A 78-year-old male with a history of hypertension and type 2 diabetes mellitus was admitted for evaluation following an episode of altered mental status and generalized weakness. On arrival, the patient was awake but only partially responsive, with a Glasgow Coma Scale score of 11 (E3 V3 M5). He was able to follow simple commands intermittently but remained disoriented to time and place.

    }}
    \caption{Example of a serialized input with text, time invariant data and time variant data}
    \label{fig:example2}
\end{figure}

This representation preserves the semantic meaning of structured variables while allowing the language model to attend to both modalities through its self-attention mechanism.

For temporal variables such as laboratory results and vital signs, measurements are first preprocessed through unit normalization and outlier filtering. The resulting values are then aggregated over the observation window before serialization.

The final serialized sequence is constructed as:

\[
x = \text{Serialize}(x_{struct}) \oplus x_{text}
\]

where $\oplus$ denotes concatenation.

\subsection{Model Architectures}

We evaluate both encoder-based and decoder-based language models using the serialized multimodal input.

\textbf{Encoder-based models.}
We employ ModernBERT as an encoder-only architecture with a context window of up to 8,192 tokens. The serialized input sequence is processed by the encoder, and the contextual representation of the \texttt{[CLS]} token is used for downstream prediction. A linear classification head is applied to produce the final output.

\textbf{Decoder-based models.}
We evaluate several generative LLM architectures, including Llama 3.1, Gemma 2B, DeepSeek-R1-Qwen 8B, and Qwen 3 8B. These models are fine-tuned using supervised instruction-style training to predict the target label as the next generated token.

In this setup, the model receives the serialized patient record as input and generates the corresponding outcome label.

\subsection{Training Objective}

For encoder-based models, training follows the standard supervised classification objective. Given the encoded representation $h_{CLS}$, the model predicts the class label using a linear layer:

\[
\hat{y} = \text{softmax}(W h_{CLS} + b)
\]

The model parameters are optimized using cross-entropy loss.

For decoder-based models, training follows the next-token prediction objective commonly used in supervised fine-tuning (SFT). The model is trained to generate the target label conditioned on the serialized patient record.

Both model types are fine-tuned on task-specific datasets using standard optimization procedures.

\section{Experimental Setup}
\label{sec:setup}

This section describes the experimental framework used to evaluate the proposed LLM-based approach against task-specific baseline models.

\subsection{Baseline Models}
We implement a diverse set of baseline architectures to reflect both state-of-the-art research methods and existing clinical infrastructure:

\begin{itemize}
    \item \textbf{In-Hospital Mortality (Unimodal \& Fusion):} We evaluate unimodal baselines using \textit{BioClinicalBERT} for clinical narratives and an \textit{LSTM} network for time-series data. For the multimodal baseline, we implement a gated fusion mechanism that combines the latent representations of both encoders.
    
    \item \textbf{Emergency Triage (Feature Concatenation):} The triage baseline utilizes a \textit{BERT} encoder for ambulance notes and a linear projection layer for vital signs, and has been introduced in more deapth in previous work \cite{maschhur2024towards}. These features are concatenated and passed through an MLP head to classify patient urgency according to the Manchester Triage System (MTS).
    
    \item \textbf{Graft Failure (Gradient Boosting):} For the transplant task, we utilize a \textbf{Gradient Boosted Decision Tree (GBDT)} model, which was used already in previous work  \cite{roller2022evaluation}. This was selected specifically because it is established in the nephology department and currently evaluated within a clinical study \cite{osmanodja2024investigating}. By using the GBDT as a baseline, we provide a direct comparison between the existing clinical standard and our proposed LLM-based approach.
\end{itemize}

\subsection{Proposed Method: Unified Textual Serialization}
Our approach avoids task-specific fusion architectures, instead converting all multimodal EHR data into a unified textual format. Structured variables such as vital signs and laboratory values are converted into textual representations and concatenated with clinical notes. This allows the language model to learn relationships across modalities using its attention mechanism. 

\begin{itemize}
    \item \textbf{Encoder-only Models:} We employ \textbf{ModernBERT} \cite{warner2025smarter} with an 8,192 token context window. The model is fine-tuned for classification using a linear prediction head applied to the \texttt{[CLS]} token representation.
    
    \item \textbf{Decoder-only Models:} We evaluate a range of generative architectures, including \textbf{Llama 3.1} \cite{grattafiori2024llama}, \textbf{Gemma 2B} \cite{team2024gemma}, \textbf{DeepSeek-R1-QWEN 8B} \cite{deepseekai2025deepseekr1}, and \textbf{QWEN 3 8B} \cite{qwen3technicalreport}. These models are trained via self-supervised fine-tuning (SFT) to predict a task-specific class token through next-token prediction.
\end{itemize}

\section{Results}
\label{sec:results}

In this section, we present the empirical evaluation of our multimodal LLM approach across three distinct clinical tasks. We evaluate our hypothesis that fine-tuned multimodal LLMs using textual serialization outperform classic fusion models and task-specific baselines.

\subsection{In-Hospital Mortality Prediction (MIMIC-III)}
The results for the ICU mortality prediction task (Table \ref{tab:mortality_results}) demonstrate that unified textual serialization effectively integrates temporal and textual information. Both the encoder-only ModernBERT and the decoder-only DeepSeek-R1-FT achieved an AUROC of 0.93 when using multimodal data (Text + TS) as input, outperforming the dedicated Text + TS Fusion baseline (0.90). Furthermore, across all evaluated models (Llama 3.1, Gemma-2-9B, and QWEN3), the multimodal configurations consistently outperformed their unimodal counterparts, confirming that the self-attention mechanisms of general-purpose LLMs successfully capture cross-modal relationships.

\begin{table}[!ht]
\begin{center}
\caption{Predictive performance for In-Hospital Mortality on the MIMIC-III dataset. (TS = Time-Series)\label{tab:mortality_results}}
\resizebox{\linewidth}{!}{%
\begin{tabular}{lccc}
\toprule
Model Configuration & AUROC & AUPR & F1-Score \\
\midrule
BioClinicalBERT (Text) & 0.83 & 0.39 & 0.40 \\
LSTM (TS) & 0.87 & 0.59 & 0.53 \\
Fusion (Text + TS) & 0.90 & 0.60 & 0.55 \\
\midrule
ModernBERT (Text) & 0.87 & 0.51 & 0.35 \\
ModernBERT (TS) & 0.89 & 0.65 & 0.40 \\
\textbf{ModernBERT (Text + TS)} & \textbf{0.93} & \textbf{0.72} & \textbf{0.62} \\
\midrule
DeepSeekR1-FT (Text) & 0.87 & 0.50 & 0.49 \\
DeepSeekR1-FT (TS) & 0.87 & 0.57 & 0.45 \\
\textbf{DeepSeekR1-FT (Text + TS)} & \textbf{0.93} & \textbf{0.72} & \textbf{0.60} \\
\midrule
Llama3.1 (Text) & 0.85 & 0.47 & 0.35 \\
Llama3.1 (TS) & 0.86 & 0.59 & 0.53 \\
Llama3.1 (Text + TS) & 0.92 & 0.69 & 0.55 \\
\midrule
Gemma-2-9B (Text) & 0.88 & 0.50 & 0.47 \\
Gemma-2-9B (TS) & 0.89 & 0.63 & 0.57 \\
Gemma-2-9B (Text + TS) & 0.91 & 0.69 & 0.59 \\
\midrule
QWEN3 (Text) & 0.88 & 0.51 & 0.49 \\
QWEN3 (TS) & 0.89 & 0.63 & 0.54 \\
QWEN3 (Text + TS) & 0.92 & 0.69 & 0.62 \\
\bottomrule
\end{tabular}%
}
\end{center}
\end{table}

\subsection{Emergency Triage Classification (KIBATIN)}
For the multi-class ambulance triage task, performance was measured using Macro, Micro, and Weighted F1-scores (Table \ref{tab:triage_results}). DeepSeek-R1 achieved the strongest overall results when all clinical features were serialized into the prompt alongside the text, yielding a Macro F1 of 0.56 compared to 0.40 for the classic fusion baseline. Interestingly, the generative models showed a strong reliance on the unstructured text modality; for example, DeepSeek-R1 achieved a 0.51 Macro F1 using text alone, compared to 0.39 using features alone. 

\begin{table}[!ht]
\begin{center}
\caption{Triage classification results (Manchester Triage System) on the KIBATIN ambulance  dataset.  \label{tab:triage_results}}
\resizebox{\linewidth}{!}{%
\begin{tabular}{lccc}
\toprule
Model & Macro F1 & Micro F1 & Weighted F1 \\
\midrule
Fusion (Text + All) & 0.40 & - & 0.57 \\
\midrule
ModernBERT (Text) & 0.42 & 0.55 & 0.52 \\
ModernBERT (Feat) & 0.43 & 0.55 & 0.50 \\
ModernBERT (Text + Feat) & 0.49 & 0.59 & 0.56 \\
\midrule
DeepSeek-R1 (Text) & 0.51 & 0.61 & 0.60 \\
DeepSeek-R1 (Feat) & 0.39 & 0.54 & 0.43 \\
\textbf{DeepSeek-R1 (Text + Feat)} & \textbf{0.56} & \textbf{0.63} & \textbf{0.63} \\
\midrule
QWEN3-9B (Text) & 0.49 & 0.59 & 0.57 \\
QWEN3-9B (Feat) & 0.39 & 0.53 & 0.40 \\
QWEN3-9B (Text + Feat) & 0.53 & 0.58 & 0.58 \\
\midrule
Gemma-2-9B (Text) & 0.48 & 0.57 & 0.54 \\
Gemma-2-9B (Feat) & 0.45 & 0.56 & 0.54 \\
Gemma-2-9B (Text + Feat) & 0.52 & 0.59 & 0.58 \\
\midrule
Llama3.2-3B (Text) & 0.46 & 0.57 & 0.55 \\
Llama3.2-3B (Feat) & 0.46 & 0.53 & 0.53 \\
Llama3.2-3B (Text + Feat) & 0.51 & 0.59 & 0.59 \\
\bottomrule
\end{tabular}%
}
\end{center}
\end{table}

\subsection{Graft Failure Prediction (PRIMA-AI)}
Table \ref{tab:graft_results} compares the proposed text-serialization models against a Gradient Boosting system deployed for kidney transplant patients. The data shows that multimodal serialization allows LLMs to successfully surpass the tabular baseline. Both ModernBERT (0.90 AUROC) and DeepSeek-R1 (0.90 AUROC) exceeded the performance of the clinical Gradient Boost model (0.89 AUROC). Furthermore, the serialized models demonstrated a noticeable improvement in AUPR (up to 0.47 vs. 0.43), indicating superior detection of the minority class (graft failure).

\begin{table}[!ht]
\begin{center}
\caption{Graft Failure prediction performance comparing LLMs to the production Gradient Boosting system.\label{tab:graft_results}}
\resizebox{\linewidth}{!}{%
\begin{tabular}{lccc}
\toprule
Model & AUROC & AUPR & F1-Score \\
\midrule
Gradient Boost (Baseline) & 0.89 & 0.43 & 0.45 \\
\midrule
ModernBERT (Text) & 0.78 & 0.26 & 0.27 \\
ModernBERT (Feat) & 0.88 & 0.42 & 0.43 \\
\textbf{ModernBERT (Text + Feat)} & \textbf{0.90} & \textbf{0.45} & \textbf{0.47} \\
\midrule
DeepSeekR1 (Text) & 0.86 & 0.36 & 0.38 \\
DeepSeekR1 (Feat) & 0.87 & 0.39 & 0.44 \\
\textbf{DeepSeekR1 (Text + Feat)} & \textbf{0.90} & \textbf{0.47} & \textbf{0.47} \\
\midrule
QWEN3-9B (Text) & 0.85 & 0.30 & 0.36 \\
QWEN3-9B (Feat) & 0.86 & 0.35 & 0.43 \\
QWEN3-9B (Text + Feat) & 0.90 & 0.47 & 0.46 \\
\midrule
Llama3.1-8B (Text) & 0.85 & 0.30 & 0.36 \\
Llama3.1-8B (Feat) & 0.88 & 0.37 & 0.42 \\
Llama3.1-8B (Text + Feat) & 0.89 & 0.47 & 0.40 \\
\midrule
Gemma2-9B (Text) & 0.86 & 0.28 & 0.36 \\
Gemma2-9B (Feat) & 0.87 & 0.39 & 0.47 \\
Gemma2-9B (Text + Feat) & 0.89 & 0.40 & 0.45 \\
\bottomrule
\end{tabular}%
}
\end{center}
\end{table}

\section{Discussion}

\textbf{Principal Findings}\\
In this study, we investigated whether large language models can act as unified multimodal learners for clinical prediction by serializing structured data into natural language sequences. Across three highly diverse clinical environments—ICU mortality, emergency ambulance triage, and outpatient graft failure prediction—our results confirm the hypothesis that fine-tuned multimodal LLMs consistently match or outperform classic task-specific fusion architectures. Most notably, standard LLM architectures fine-tuned on serialized text representations outperformed a production-grade Gradient Boosted Decision Tree (GBDT) model in predicting long-term graft failure, highlighting the clinical viability of this paradigm.

\textbf{Architectural Insights: Encoders vs. Decoders}\\
Our evaluation provides a rigorous comparison between encoder-only (ModernBERT) and decoder-only (e.g., DeepSeek-R1, QWEN3) architectures. The data reveals that both model families are highly capable of mapping cross-modal relationships through self-attention. In the MIMIC-III mortality task and the PRIMA-AI graft failure task, ModernBERT effectively synthesized the disparate modalities, showing clear performance gains when text and structured features were combined (e.g., jumping from 0.88 to 0.90 AUROC in graft prediction). 

However, generative decoder models demonstrated a particular robustness in noisy, unstructured environments. In the KIBATIN triage task, which relies heavily on short, abbreviated ambulance notes, generative models like DeepSeek-R1 (Macro F1 0.56) significantly outperformed both the fusion baseline (0.40) and the encoder-based ModernBERT (0.49). We hypothesize that decoder-only models, pre-trained on vastly larger and more diverse corpora, possess superior robustness when interpreting the sparse text typical of emergency settings.

\textbf{Clinical and System Implications}\\
The prevailing paradigm of task-specific fusion creates significant friction in deploying clinical decision support systems. Every new task typically requires custom preprocessing pipelines, isolated encoders, and specific hyperparameter tuning. By demonstrating that a single, unified text serialization pipeline can simultaneously process time-series data, tabular demographics, and unstructured clinical notes, we offer a pathway to drastically simplify clinical machine learning infrastructure.

\textbf{Limitations and Future Work}\\
Despite its empirical success, text serialization naturally expands the dimensionality of structured data. A numeric laboratory matrix that requires minimal memory in a traditional tabular model consumes many more tokens when serialized, which may challenge context windows for patients with massive longitudinal histories. Future work should investigate efficient serialization strategies, as well as multimodal interpretability to ensure these models can provide clinician-trusted rationales for their predictions.

\section{Conclusion}

Integrating heterogeneous clinical modalities has historically required complex, bespoke fusion architectures. In this study, we demonstrated that standard Large Language Models can serve as unified, high-performing multimodal learners simply by serializing structured clinical data into natural language sequences. Our comprehensive evaluation across critical care, emergency triage, and transplant nephrology shows that this unified approach consistently outperforms both specialized fusion models and traditional tabular ML baselines. By unifying time-series, tabular features, and text into a single representational space, this paradigm significantly reduces the engineering overhead of clinical decision support systems while maximizing predictive accuracy.

\section*{Limitations}



\bibliography{custom}

\appendix

\section{Examples: Instruction Tuning Prompts}
\label{sec:appendix}

Figure \ref{fig:example_prompt} presents an example prompt used for instruction tuning to forecast graft failure.

\begin{figure*}[ht]
    \centering
    \small
    \fbox{\parbox{0.95\textwidth}{
Du bist ein Klassifikationsmodell zur Bestimmung des Risikos eines Transplantatversagens in der Nephrologie.
Dir stehen Patientendaten aus verschiedenen Quellen wie klinischen Messwerten, Laborbefunden und Arztberichten zur Verfügung.\\
\\
\textbf{\#\#\# Demografische Angaben}:\\
Alter: 54\\
Geschlecht: männlich\\
Blutgruppe: A-\\
 \\
\textbf{\#\#\# Lebensstil und Risikofaktoren}:\\
Raucher: Nein\\
Alkoholismus: Nein\\
fam. Hypertonus: Nein\\
Medikamentenabu: Nein\\
Analalgetikaabu: Nein\\
Hyperlipoprotei: Nein\\
 \\
\textbf{\#\#\# Klinische Messungen}:\\
Blutdruck systolisch: 119.0, 127.0\\
Blutdruck diastolisch: 83.0, 87.0\\
Herzfrequenz: 112.0, 123.0\\
Temperatur: NA, NA\\
Gewicht: 70.0, 70.5\\
Urinvolumen: NA, 2400.0\\
Diuresezeit: 24.0, 24.0\\
 \\
\textbf{\#\#\# Laborwerte}:\\
Labor Kreatinin: NA, 2.63\\
Labor Urea: NA, 86.0\\
Labor Albuminurine: NA, NA\\
Labor Proteinurine: NA, NA\\
Labor CRP: NA, 7.16\\
Labor Eryeb: NA, 4.44\\
Labor Leukourine: NA, 1.0\\
Labor HB: NA, 12.6\\
Labor Stabicarb: NA, NA\\
 \\
\textbf{\#\#\# Transplantations- und Verlaufsdaten}:\\
Transplantation Nummer Tx insgesamt: 15\\
Transplantation Spenderart: 0\\
Transplantation HbS AG: 0\\
Transplantation HCV AK: 0\\
Transplantation CMV AK: 0\\
Transplantation Dialyseart: HD\\
Transplantation Ischaemie kalt: 14.0\\
Transplantation Primaerfunktion: 1\\
Transplantation MM Broad: 3.0\\
Transplantation MM Split: 1.0\\
Monate seit Transplantation: 18\\
Anzahl an Aufenthalten im Krankenhaus: 12\\
 \\
\textbf{\#\#\# Ärztlicher Befund}:\\
Beurteilung: Tx Funktion stabil. Stationäre Aufnahme auf 124 wie besprochen. Pat. geht es nicht so gut, ist seit letzter Woche heiser. Pat. wirkt etwas verändert im Verhalten. Pat. hat seit 27.09.2023 selbständig Valcyte trotz bestehender Virusaktivität und im therapeutischen Bereich liegender Ganciclovirspiegel abgesetzt. Jetzt Virusbefund abwarten. Berichtet Taubheitsgefühle in Fingern, Zunge, Lippe und Waden seit letzter Woche NA> könnten NW von Valcyte sein. Kein Fieber, sonst keine Beschwerden. Keine Dyspnoe. Appetit gut. ...\\
 \\
Basierend auf den obigen Informationen, klassifizieren Sie, ob der Patient innerhalb der nächsten 360 Tage ein Transplantatversagen erleiden wird.\\
\textbf{class 1}: Nein (kein Transplantatversagen erwartet)\\
\textbf{class 2}: Ja (Transplantatversagen erwartet)\\
\\
\textbf{\#\#\# LÖSUNG}:\\
Die korrekte Antwort ist: class 1\\

    }}
    \caption{German prompt use for instruction tuning to predict graft failure.}
    \label{fig:example_prompt}
\end{figure*}

\end{document}